\documentclass[a4paper]{spie}  

\usepackage{amsmath,amsfonts,amssymb}
\usepackage{graphicx}
\usepackage{booktabs}
\usepackage{array}
\usepackage[table]{xcolor}
\usepackage[colorlinks=true, allcolors=blue]{hyperref}
\usepackage{multirow}
\usepackage{svg}

\title{Cross-modal learning for SAR target recognition using optical vision foundation models}

\author[a]{Lucas Hirsch\textsuperscript{*}}
\author[a]{James R. Hopgood}
\author[b]{Javid Khan}
\author[c]{Yoann Altmann}
\author[a]{Mike Davies\textsuperscript{*}}
\affil[a]{School of Engineering, University of Edinburgh, United Kingdom}
\affil[b]{Leonardo UK, United Kingdom}
\affil[c]{School of Engineering and Physical Sciences, Heriot-Watt University, United Kingdom}

\authorinfo{\textsuperscript{*}Corresponding authors:
Lucas Hirsch (Lucas.Hirsch@ed.ac.uk) and
Mike Davies (Mike.Davies@ed.ac.uk).}

\begin{document} 
\maketitle

\begin{abstract}

Synthetic Aperture Radar (SAR) is an important modality in a wide range of imaging applications due to its versatile, long range and near all weather operating capabilities. However, Automatic Target Recognition (ATR) remains a challenging problem due to limited labelled data, the strong speckle in SAR images and the significant domain gap between SAR and more abundant optical imagery. In contrast, electro-optical (EO) imagery benefits from massive datasets, clearer visual structure and powerful foundation models. In this work, we investigate how vision foundation models trained on optical data can provide class level supervision for SAR classification.

We propose a cross-modal EO to SAR prototype alignment framework in which a frozen EO encoder, based on a DINOv3 vision foundation model, is used to construct class level optical prototypes without requiring strict EO/SAR pairs. A SAR model is then trained to classify SAR images while aligning its embeddings to the corresponding EO class prototype. At inference time, the SAR model operates independently, without access to optical imagery.
We evaluate our approach on the UNICORNv2 dataset, an EO and SAR dataset of civilian vehicles with heavily speckled images and severe class imbalance. EO prototype alignment improves SAR classification accuracy over frozen DINOv3, SAR only finetuning and unpaired distribution alignment baselines, and t-SNE visualizations provide qualitative evidence of clearer separation among classes in the trained SAR embedding space. These results suggest that optical vision foundation models, despite being trained on visible spectrum imagery, provide transferable information for SAR image classification, offering a practical method for using large scale pretrained vision foundation models across challenging sensing modalities.

\end{abstract}

\keywords{Synthetic aperture radar, automatic target recognition, cross-modal learning, multimodal learning, domain adaptation, foundation models, transfer learning} 

\section{INTRODUCTION}
\label{sec:intro}



Synthetic Aperture Radar (SAR) is a valuable sensing modality for remote observation due to its capability to image over long ranges and under operating conditions that are challenging for passive optical sensors. 
These properties make SAR useful for automatic target recognition (ATR) in defence and autonomous sensing applications. Still, SAR ATR remains difficult. Images are heavily affected by speckle, less visually interpretable than electro-optical (EO) images and radar returns are dependent on target geometry and viewing angle\cite{cruz2022_reviewSAR_imageFormation, frascatoro21_reviewSAR_despeckling}.
In addition, labelled SAR target datasets are typically smaller and cover a narrower range of target classes than the optical datasets used to train modern vision systems \cite{liu2026_atrnet,ramos2026mavic,simeoni2025dinov3}.

In contrast, optical imagery benefits from large existing datasets and more visually interpretable structure, which in recent years has allowed for the training of powerful self-supervised vision foundation models. DINOv3, the latest vision foundation model by Meta AI\cite{simeoni2025dinov3}, was trained on 1.7 billion images and produces transferable visual representations across a wide range of tasks, such as object detection, segmentation and classification. 

The strength of these foundation models in analyzing optical imagery poses an interesting question for SAR ATR: can features learned by an optical foundation model provide transferable information for SAR classification, even when optical imagery is not available at inference time? Although SAR and EO images are produced by different sensing mechanisms, target shape and geometry can provide shared structure across the two modalities. This suggests that optical foundation models may encode information that can be transferred to a SAR model during training.

A straightforward way to use EO data for SAR ATR is to rely on coincident EO/SAR observations during training \cite{ramos2026mavic, liu2025_MAVICEntry, ma2025sarclip}. This is useful when paired observations are available, but requiring one-to-one sample correspondence between modalities is a strong assumption. In practice, it would be more realistic to have labelled EO examples from the same target classes, possibly collected at a different time or
drawn from a separate archive, rather than requiring each SAR image to have a matched EO counterpart from the same acquisition event.

This work therefore uses class level EO references rather than sample level EO/SAR pairs. EO images are passed through a strong optical foundation model and averaged within each class to construct ``prototype'' vectors\cite{snell17_prototypicalNetworks}. Each prototype is a single vector that represents an entire class (such as a sedan, bus or truck), and captures its relevant information in the optical feature space. A SAR model is then trained to classify SAR chips while aligning its embeddings to the corresponding EO class prototype. At inference time, the EO branch is discarded and only the SAR classifier is retained.
The aim is to use the optical model's ability to extract strong representations from the EO imagery and transfer part of this to the SAR model through cross-modal training.

The contribution of this paper is a cross-modal prototype alignment framework for SAR target classification. A frozen EO DINOv3 encoder is used to construct class prototypes, while a SAR DINOv3 encoder is adapted using Low-Rank Adaptation (LoRA)\cite{hu2022LoRA}. The SAR model is trained with a classification loss and a prototype alignment loss, producing a SAR classifier whose feature space is encouraged to follow class structure encoded by the EO prototypes. The experimental evaluation uses a challenging EO/SAR vehicle recognition benchmark with strong class imbalance, low resolution chips and severe SAR speckle\cite{leong2019_unicorn_v1_dataset,low2023mavic}.
We compare a frozen DINOv3 baseline, a baseline finetuned using SAR alone and an unpaired Maximum Mean Discrepancy (MMD) alignment baseline. Results show that EO prototype alignment achieves the highest mean top-1 accuracy and macro-F1 score among the evaluated methods. A control experiment using uniformly separated synthetic prototypes suggests that the improvement is not exclusively explained by latent space regularization, but also by class structure transferred from the EO feature space.

\section{RELATED WORK}
\label{sec:related}

\subsection{SAR target recognition}
Synthetic Aperture Radar Automatic Target Recognition (SAR ATR) aims to classify targets of interest in SAR imagery.
Deep learning has become a common approach for SAR ATR\cite{zhou2025_FiftyYearsSarATR}, but SAR imagery poses different challenges from natural or EO imagery. The same object class can produce different returns depending on viewing angle, radar scattering effects and image processing conditions, while visually distinct object classes may have overlapping SAR signatures\cite{lang2025_recentAdvancesSarATR}. These factors are particularly problematic for small target chips and imbalanced datasets where minority classes are poorly represented during training \cite{ramos2026mavic}. These challenges motivate the use of transfer learning and information from other sensing modalities.

\subsection{Cross-modal SAR and EO learning}
Typically, EO and SAR sensors provide complementary information. EO images contain more interpretable shape and texture, while SAR captures radar scattering properties and can operate in adverse illumination or weather conditions. Although the modalities are physically different, some structure can be shared. For example, rough object shape and strong reflective edges may be visible in both modalities, but can be difficult to interpret in SAR because of speckle or radar specific image formation effects\cite{liu2026_atrnet}. This motivates the use of EO imagery as a source of complementary information for SAR representation learning.

Prior multimodal aerial recognition work has used paired EO/SAR data to learn shared or aligned representations \cite{low2023mavic, liu2025_MAVICEntry, mishra2026_crossmodal_aerial_wildfire}. A relevant example is the annual Multi-modal Aerial View Imagery Challenge: Classification (MAVIC-C), run as part of the Perception Beyond the Visible Spectrum CVPR Workshop, which introduced the UNICORNv2 dataset used in this study\cite{low2023mavic,ramos2026mavic}. MAVIC-C considers vehicle recognition using coincident EO/SAR data for training and SAR imagery only for evaluation. Several challenge submissions use the correspondence between EO and SAR samples explicitly in their training objectives\cite{low2023mavic,ramos2026mavic,liu2025_MAVICEntry}. 
%

In contrast, the proposed method uses a less restrictive form of cross-modal supervision. Although the dataset contains coincident EO/SAR chips, the method does not use their pair correspondences. Instead, EO data is aggregated through class-level prototypes, so training requires shared class labels but not matched EO/SAR observations.

\subsection{Foundation models and finetuning}
Vision foundation models provide strong generic representations, but their direct application to SAR is limited by the differences between optical and radar image formation. Recent research has developed foundation models designed specifically for SAR\cite{li25_SARATR-X} and SAR vision-language models \cite{ma2025sarclip}. However, SAR pretraining datasets at the scale used for optical foundation model training are much less readily available. Other work has instead
adapted optical foundation models to SAR, with results showing that they can outperform SAR foundation models\cite{Inkawhich2025_FoundationATR}.

Our proposed approach is complementary to direct finetuning. Instead of using an optical foundation model only as a backbone for SAR finetuning, we also use a frozen EO encoder as a source of class level reference features for alignment. The SAR encoder is then adapted using Low-Rank Adaptation (LoRA), which freezes the pretrained backbone and introduces trainable low-rank updates\cite{hu2022LoRA}.


\section{METHOD}
\label{sec:method}

\begin{figure}[htbp]
\centering
\includegraphics[width=0.9\textwidth]{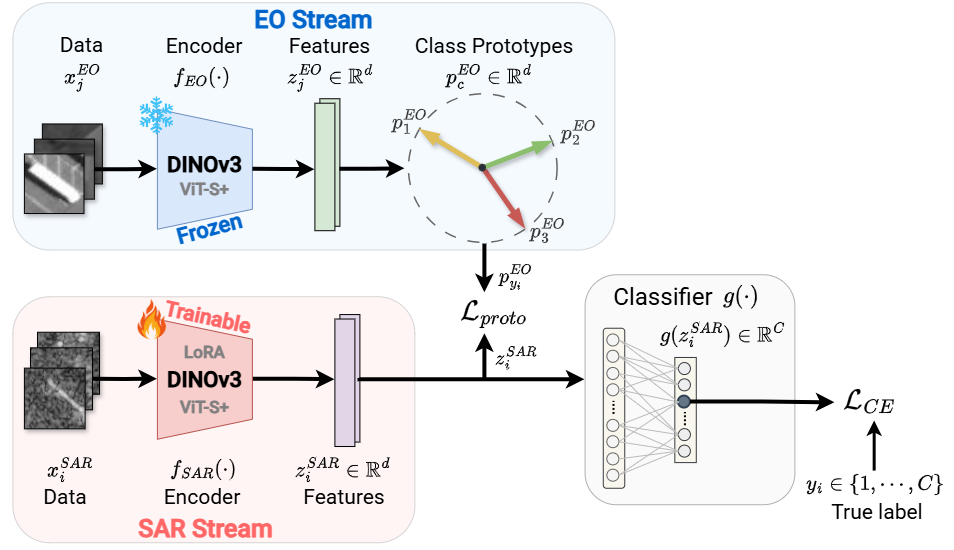}
\caption{Overview of the proposed cross-modal prototype alignment framework. The EO branch is used only to construct class level reference prototypes used during training. The prototype loss $\mathcal{L}_{proto}$ aligns the SAR features to the EO class prototypes and $\mathcal{L}_{CE}$ is the cross-entropy loss used for SAR classification. At inference time, only the SAR branch is used.}
\label{fig:method}
\end{figure}

\subsection{Problem setting and overview}
Let $\mathcal{D}_{SAR}=\{(x_i^{SAR},y_i)\}_{i=1}^{N_S}$ denote the set of labelled SAR training chips from $C$ target classes, with $x^{SAR}_i\in\mathbb{R}^{n\times n}$ and $y_i \in \{1,\ldots,C\}$.
We also assume access to labelled EO examples $\mathcal{D}_{EO}=\{(x_j^{EO},y_j)\}_{j=1}^{N_E}$ from the same class set. The goal is to train a classifier that predicts $\hat{y}_i$ from the SAR image $x_i^{SAR}$ alone, where $\hat{y}_i$ denotes the predicted class and $y_i$ denotes the true class. The EO data is used during training, but is discarded at inference. 

We use a cross-modal prototype alignment framework in which a frozen EO encoder provides class level prototype targets, and a SAR encoder is trained to classify SAR chips while aligning its embeddings to those targets. The proposed architecture is shown in Fig.~\ref{fig:method}. The EO branch uses a frozen DINOv3 encoder to compute prototypes $p^{EO}_c\in\mathbb{R}^d$ for each of the $C$ target classes. These prototypes are precomputed and used only during training. The SAR branch processes the SAR imagery with a separate DINOv3 encoder $f_{SAR}(\cdot)$, finetuned via LoRA. The resulting SAR embeddings $z_i^{SAR}$ are passed to a linear classifier $g(\cdot)$ to predict one of the $C$ classes. Two loss functions are used to guide the training: a standard cross-entropy loss $\mathcal{L}_{CE}$ for classification and a prototype alignment loss $\mathcal{L}_{proto}$  that encourages SAR embeddings to align with the EO prototype of their class.
At inference, only the SAR branch is used for prediction. 

The method does not require correspondence between individual SAR and EO samples. Instead, EO examples are aggregated at the class level to create prototypes\cite{snell17_prototypicalNetworks}, so an EO image used to construct a prototype does not need to be paired with any particular SAR training chip. In principle, the EO data could come from a separate archive collected at a different time, provided that it contains the same target classes and broadly comparable object viewpoints. Optical foundation model features are designed to be robust to variations such as small translations, crops and scale changes, but the EO examples should still represent the relevant target appearances.

\subsection{EO prototype construction}
The EO stream uses a frozen DINOv3 ViT-S+ encoder $f_{EO}(\cdot)$ trained on optical imagery, with embedding dimension $d = 384$. For each EO image, we extract and normalize its feature vector as 
\begin{equation}
    z_j^{EO}
    =
    \frac{f_{EO}(x_j^{EO})}
    {\|f_{EO}(x_j^{EO})\|_2}
    \in \mathbb{R}^d.
\end{equation}
For each class $c$, let $\mathcal{I}_c^{EO}$ denote the set of EO training indices belonging to that class
\begin{equation}
    \mathcal{I}_c^{EO}=\{j:y_j=c\},
\end{equation}
with $N_c=|\mathcal{I}_c^{EO}|$ the number of EO examples in class $c$. 
The corresponding EO prototype is computed by averaging the normalized features within the class and normalizing the result
\begin{equation}
    \tilde{p}^{EO}_c
    =
    \frac{1}{N_c}
    \sum_{j\in\mathcal{I}_c^{EO}} z_j^{EO},
    \qquad
    p^{EO}_c
    =
    \frac{\tilde{p}^{EO}_c}
    {\|\tilde{p}^{EO}_c\|_2}.
    \label{eq:prototype}
\end{equation}
Each prototype therefore has unit norm and provides a class reference in the EO feature space. Because they are computed by averaging over all the EO examples in a class, this step removes pair identities and does not require the matched EO counterpart for each SAR training example.


\subsection{SAR encoder and classification head}
The SAR branch uses a second DINOv3 ViT-S+ encoder $f_{SAR}(\cdot)$ with LoRA adapters for training. The base model weights remain frozen and only the LoRA parameters are updated. For a SAR image $x_i^{SAR}$, the normalized embedding $z^{SAR}_i\in\mathbb{R}^d$ is
\begin{equation}
    z^{SAR}_i = \frac{f_{SAR}(x_i^{SAR})}{\|f_{SAR}(x_i^{SAR})\|_2}.
    \label{eq:sar_embedding}
\end{equation}
A linear classifier $g(\cdot)$ maps SAR embeddings to class logits $g(z_i^{SAR}) \in \mathbb{R}^C$.
Classification is trained with the standard cross-entropy loss over a batch of size $B$:
\begin{equation}
    \mathcal{L}_{CE} =  -\frac{1}{B}\sum_{i=1}^{B} \log \left[\mathrm{softmax}(g(z_i^{SAR}))_{y_i}\right],
    \label{eq:cross_entropy}
\end{equation}
where $\mathrm{softmax}(g(z_i^{SAR}))_{y_i}$ denotes the predicted probability
assigned to the true class $y_i$.

\subsection{Prototype alignment loss}
The prototype alignment loss encourages each SAR embedding to align with the EO
prototype of its class. Since both $z_i^{SAR}$ and $p_{y_i}^{EO}$ are normalized, their inner product is the cosine similarity.
For a batch of size $B$ we define
\begin{equation}
    \mathcal{L}_{proto} = \frac{1}{B}\sum_{i=1}^{B} \left(1 - \langle z_i^{SAR}, p^{EO}_{y_i} \rangle\right),
    \label{eq:proto_loss}
\end{equation}
where  $p^{EO}_{y_i}$ is the EO prototype corresponding to the class label $y_i$ and $\langle \cdot, \cdot \rangle$ is the inner product. When the vectors point in the same direction, their inner product equals one and the corresponding loss equals zero. As the angle between them increases, the inner product decreases and the loss increases.
This objective therefore pulls SAR embeddings towards the prototypes constructed from the EO data.

The complete training objective combines the classification loss \eqref{eq:cross_entropy} and the prototype alignment loss \eqref{eq:proto_loss}
\begin{equation}
    \mathcal{L} = (1-\lambda)\mathcal{L}_{CE} + \lambda \mathcal{L}_{proto},
    \label{eq:full_loss}
\end{equation}
where $\lambda$ controls the contribution of each term.

\begin{figure}[htbp]
\centering
\includegraphics[width=0.7\textwidth]{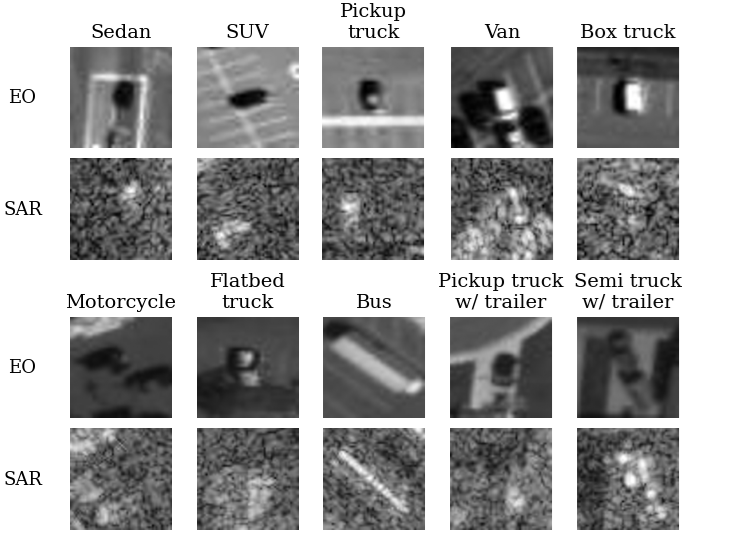}
\caption{Examples of SAR and EO vehicle chips in the UNICORNv2 dataset.}
\label{fig:dataset}
\end{figure}

\subsection{Baselines}

We compare the proposed EO prototype alignment method against three baselines:

\begin{itemize}
    \item \textbf{Frozen DINOv3:} a pretrained DINOv3 encoder is applied to SAR images and kept frozen, while only a linear classifier is trained.

    \item \textbf{SAR only finetuning:} a DINOv3 encoder is adapted with LoRA using only a cross-entropy loss and SAR data.

    \item \textbf{Unpaired MMD alignment:} global SAR and EO feature distributions are aligned using Maximum Mean Discrepancy (MMD) distance, without class level prototypes. MMD is a widely used domain adaptation approach that measures a kernel based distance between two feature distributions \cite{long15_domainAdaptation}. Minimizing this loss encourages SAR and EO embeddings to have similar global statistics, rather than matching individual samples or class prototypes.
\end{itemize}

Together, these baselines separate the effects of SAR finetuning, generic cross-modal distribution alignment and EO prototype supervision.

\section{EXPERIMENTAL SETUP}
\label{sec:setup}

\subsection{Dataset}
Experiments are conducted on UNICORNv2, a refined version of the UNIfied COincident Optical and Radar for recognitioN (UNICORN) 2008 dataset \cite{leong2019_unicorn_v1_dataset}. UNICORNv2 
contains Wide Area Motion Imagery (WAMI) EO sensor data and Wide Area SAR data collected from an aircraft over Dayton, Ohio \cite{ramos2026mavic}. The original collection used an airborne six-camera monochrome WAMI EO system and a modified GOTCHA/NDU SAR sensor. Both sensor streams operated at approximately 2~Hz, with the SAR sensor covering a fixed $5\times5$~km area that overlapped the EO field of view. The original GOTCHA system operated at X-band, but the operating band of the modified UNICORN radar is not explicitly documented\cite{Casteel2007ACP}.
Relative to the original UNICORN release, UNICORNv2 improves alignment accuracy and increases the number of labelled image chips.

The target set contains ten vehicle classes: sedan, SUV, pickup truck, van, box truck, motorcycle, flatbed truck, bus, pickup truck with trailer, and semi truck with trailer. The dataset contains approximately $450,000$ images per modality, with each object having a corresponding SAR and EO chip. Examples are shown in Fig.~\ref{fig:dataset}. The EO images clearly show the vehicles, while the SAR chips suffer from heavy speckle. Some images may contain more than one vehicle, but the labeled object of interest is usually centered. The dataset is also highly imbalanced, with common vehicle types appearing much more frequently than rare classes, as shown in Table \ref{tab:unicornv2_class_distribution}. This class imbalance is important because an improvement in overall accuracy can still hide poor behaviour on minority categories.

\begin{table}[!h]
\centering
\begin{tabular}{@{}lrr@{}}
\toprule
Vehicle type & \# Train & \# Test \\
\midrule
Sedan & 364,291 & 77 \\
SUV & 43,401 & 77 \\
Pickup truck & 24,158 & 77 \\
Van & 16,890 & 77 \\
Box truck & 2,896 & 77 \\
Motorcycle & 1,441 & 77 \\
Flatbed truck & 898 & 77 \\
Bus & 612 & 77 \\
Pickup truck w/ trailer & 695 & 77 \\
Semi truck w/ trailer & 353 & 77 \\
\midrule
\textbf{Total} & \textbf{455,635} & \textbf{770} \\
\bottomrule
\end{tabular}
\vspace{4pt}
\caption{Class distribution for the UNICORNv2 SAR training and testing splits. The test set is balanced, while the training set presents 1000:1 imbalance.}
\label{tab:unicornv2_class_distribution}
\end{table}

In addition to evaluating classification over the original ten classes, we also report performance on a coarser seven class problem, in which sedans, SUVs, pickup trucks and vans are merged into a single \emph{light vehicle} superclass. This grouping is motivated by the similar SAR signatures these vehicles have and the analysis presented in Section~\ref{sec:tsne}. The models remain trained to distinguish all ten classes, the merge is applied only when computing accuracy over the resulting seven classes.


\subsection{Implementation and evaluation details}
Both EO and SAR images are resized to $128\times128$ and normalized using ImageNet statistics. The EO DINOv3 encoder is frozen and used only to precompute class prototypes. The SAR DINOv3 encoder is adapted with LoRA and trained with a linear classifier head. Further implementation details and a link to the public GitHub repository can be found in Appendix~\ref{sec:appendix}. 

For the results, we report top-1 classification accuracy and macro-F1.
Given $N$ test samples, top-1 accuracy is defined as
\begin{equation}
    \mathrm{Accuracy}
    = \frac{1}{N}\sum_{i=1}^{N}
    \mathbf{1}\!\left(\hat{y}_i = y_i\right),
\end{equation}
where $y_i$ and $\hat{y}_i$ are the true and predicted class labels and $\mathbf{1}\!\left(\cdot\right)$ is the indicator function.
Macro-F1 is the average of the class F1 scores
\begin{equation}
    \mathrm{Macro\text{-}F1}
    = \frac{1}{C}\sum_{c=1}^{C} F1_c =\frac{1}{C}\sum_{c=1}^{C}
    \frac{2\,\mathrm{TP}_c}
    {2\,\mathrm{TP}_c+\mathrm{FP}_c+\mathrm{FN}_c},
\end{equation}
where $C$ is the number of classes, and $\mathrm{TP}_c$, $\mathrm{FP}_c$ and $\mathrm{FN}_c$ denote the true positives, false positives and false negatives for class $c$, respectively. Macro-F1 gives equal weight to each class and complements overall accuracy by showing whether performance is distributed consistently across the classes. Reported accuracy results are averaged over five repeated runs and uncertainty is reported as one standard deviation.

\section{RESULTS}
\label{sec:results}

\subsection{Classification performance}
Table~\ref{tab:main_results} reports classification performance under both evaluation settings. For the evaluation over the original ten classes, the frozen DINOv3 baseline reaches $26.9\%$ accuracy, showing that optical foundation model features are not directly sufficient for SAR ATR without adaptation. LoRA finetuning improves accuracy to $29.8\%$, but gives only a small macro-F1 improvement, suggesting that finetuning using SAR alone continues to struggle with the imbalance in the training data. Unpaired MMD alignment improves both metrics, reaching $31.8\%$ accuracy and $0.277$ macro-F1. The proposed EO prototype alignment achieves the highest mean performance, reaching $33.3\%$  accuracy and $0.281$ macro-F1.

\begin{table}[htbp]
\centering
\begin{tabular}{@{}lccc@{}}
\toprule
Method & Evaluation & Acc. (\%) & Macro-F1 \\
\midrule
Frozen DINOv3
    & \multirow{4}{*}{10 classes}
    & $26.9 \pm 0.2$ & 0.235 \\
DINOv3 LoRA finetuning
    &
    & $29.8 \pm 1.4$ & 0.244 \\
Unpaired MMD alignment
    &
    & $31.8 \pm 1.6$ & 0.277 \\
EO prototype alignment
    &
    & $\mathbf{33.3 \pm 1.3}$ & $\mathbf{0.281}$ \\
\midrule
\midrule
Frozen DINOv3
    & \multirow{2}{*}{7 classes}
    & $41.1 \pm 1.1$ & 0.305 \\
EO prototype alignment
    &
    & $\mathbf{52.2 \pm 1.3}$ & $\mathbf{0.384}$ \\
\bottomrule
\end{tabular}
\vspace{4pt}
\caption{SAR classification performance on UNICORNv2. The seven class evaluation merges sedans, SUVs, pickup trucks and vans into one \emph{light vehicle} superclass. Results are averaged across five runs, with accuracy reported as mean $\pm$ standard deviation. Bold values indicate the best result within each evaluation.}
\label{tab:main_results}
\end{table}

Within the ten class evaluation, the gain from EO prototype alignment is modest in absolute terms but consistent with the difficulty of the problem. Relative to the frozen DINOv3 baseline, prototype alignment improves accuracy by $6.4$ percentage points. Relative to LoRA finetuning using SAR alone, it improves accuracy by $3.5$ percentage points. Both cross-modal alignment methods, MMD and EO prototype alignment, outperform frozen DINOv3 and LoRA finetuning using SAR alone, supporting the central claim that EO references contain transferable information for SAR representation learning. 

The lower block of Table~\ref{tab:main_results} reports a focused comparison for the coarser evaluation over seven classes defined in Section~\ref{sec:setup}. After merging sedans, SUVs, pickup trucks and vans into the light vehicle category, EO prototype alignment improves accuracy from $41.1\%$ to $52.2\%$ and macro-F1 from $0.305$ to $0.384$ relative to frozen DINOv3. This suggests that prototype alignment improves coarse vehicle recognition even when closely related vehicle classes cannot be reliably distinguished.

One possible explanation for the higher performance of prototype alignment over MMD is that it provides a more direct training signal. MMD reduces the discrepancy between the EO and SAR feature distributions, but does not directly specify where each SAR embedding should lie since it works at a distribution level. In contrast, prototype alignment pulls each SAR embedding towards a fixed EO reference for its class.

\subsection{Feature space visualization}
\label{sec:tsne}
\begin{figure}[htbp]
\centering
\includegraphics[width=1\textwidth]{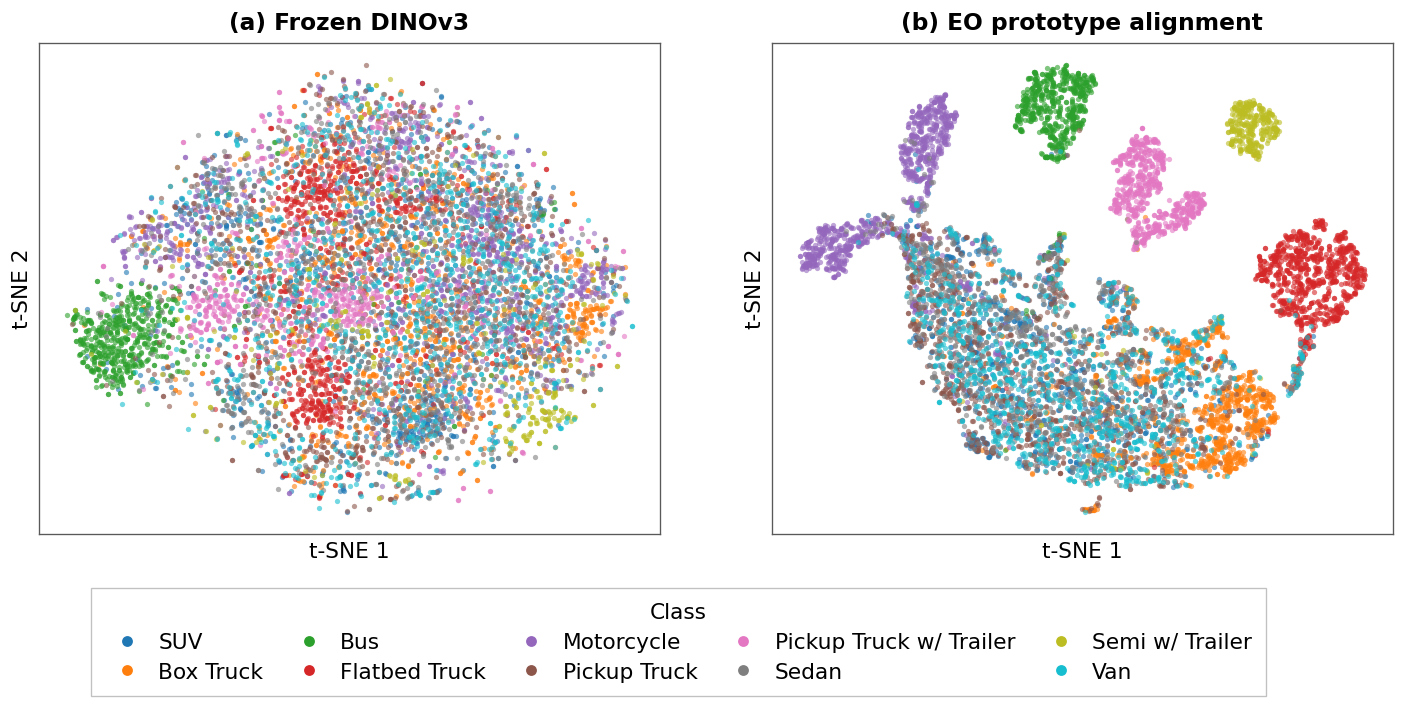}
\caption{Feature space visualization with t-SNE of the SAR embeddings for the frozen DINOv3 baseline (left) and the prototype aligned model (right). Colours indicate vehicle classes. The prototype aligned embeddings appear better separated in the two dimensional projection.}
\label{fig:tsne}
\end{figure}

To complement the quantitative results, we visualize the SAR feature space using t-SNE\cite{vandermaaten2008_tsne}. We compare embeddings from the frozen DINOv3 baseline with those from the prototype aligned model, using up to 1000 training samples from each class. The results are shown in Fig.~\ref{fig:tsne}, with the frozen DINOv3 baseline on the left and EO prototype alignment on the right.

The frozen DINOv3 embeddings in Fig.~\ref{fig:tsne} show overlap between most classes, indicating that the optical foundation model features do not separate the SAR data cleanly without adaptation. Buses are the clearest exception, possibly due to their distinct size and SAR signature. After EO prototype alignment, the t-SNE projection shows more separated clusters, broadly consistent with the improvement in accuracy and macro-F1 reported in Table~\ref{tab:main_results}. Classes with distinct shape or scale such as buses, box trucks and pickup trucks with trailers, form clear clusters, while sedans, SUVs, vans and pickup trucks (the light vehicles) continue to overlap in the central region. This overlap, together with their similar SAR signatures and the classification errors among these classes, motivates the light vehicle superclass evaluation reported in the lower section of Table~\ref{tab:main_results}.


To examine where the gains from EO prototype alignment arise, we use Table~\ref{tab:per_class_recall} which reports the per-class accuracy over all the runs. The frozen DINOv3 model achieves its highest recall for buses, pickup trucks with trailers and sedans. EO prototype alignment improves recall for box trucks, SUVs, buses, pickup trucks and vans, while pickup trucks with trailers remain among the most reliably identified classes. The improvements are not uniform, with a decrease in sedan recall and a persistent difficulty in recovering flatbed trucks, motorcycles and semi trucks with trailers.
The decrease in sedan recall may be related to sedans accounting for most of the training examples. The baseline may therefore overpredict this class when presented with an ambiguous signal. Prototype alignment appears to redistribute some of these predictions among the other light vehicle classes. 

Motorcycles, flatbed trucks and semi trucks with trailers appear clustered in the t-SNE projection, but remain poorly classified by both models. This discrepancy shows an important limitation of the visualization. The projection uses training embeddings, whereas classification is evaluated on the test set. Moreover, t-SNE is a nonlinear projection that prioritizes local neighbourhoods and can exaggerate gaps between clusters \cite{vandermaaten2008_tsne}. The plots should therefore be interpreted qualitatively rather than as clear evidence of linear separability in the original feature space.


\begin{table}[htbp]
\centering
\begin{tabular}{@{}lcc@{}}
\toprule
\multicolumn{3}{c}{\textbf{Per-class accuracy (\%)}} \\
\midrule
Class & Frozen DINOv3 & EO prototypes \\
\midrule
SUV                       & 7.6           & \textbf{17.8} \\
Box truck                 & 25.8          & \textbf{61.5} \\
Bus                       & 70.9          & \textbf{87.0} \\
Flatbed truck             & 0.0           & 0.0 \\
Motorcycle                & \textbf{6.9}  & 3.3  \\
Pickup truck              & 13.6          & \textbf{29.3} \\
Pickup truck with trailer & 86.1          & \textbf{90.0} \\
Sedan                     & \textbf{41.6} & 19.2 \\
Semi truck with trailer   & \textbf{3.4}  & 0.0 \\
Van                       & 13.5          & \textbf{28.7} \\
\bottomrule
\end{tabular}
\vspace{4pt}
\caption{Mean per-class accuracy across five runs for frozen DINOv3 and EO prototype alignment. Per-class accuracy is the proportion of examples from each class that are correctly classified}
\label{tab:per_class_recall}
\end{table}

\subsection{Is EO alignment just regularization?}
\label{sec:EO_regularization}
\begin{figure}[htbp]
\centering
\includegraphics[width=0.65\textwidth,     
                trim={0.1cm 0.45cm 0.1cm 0.45cm},
                clip]{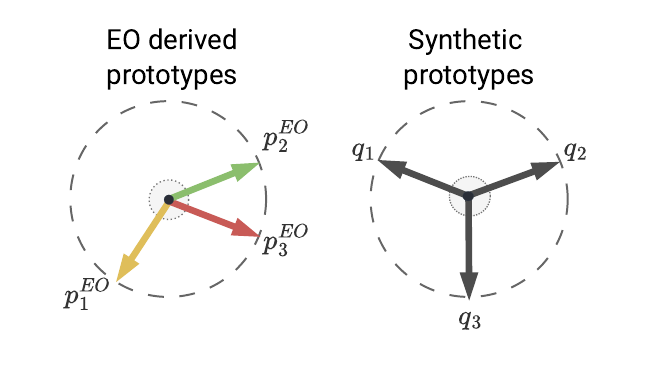}
\caption{Schematic comparison of EO ($p_i^{EO}$) and synthetic ($q_i$) prototypes. Three unit vectors are shown in two dimensions for illustration. EO prototypes exhibit geometry learned from optical imagery, with class relationships represented by their pairwise angles. Synthetic prototypes are uniformly separated and contain no information from EO data. The experiments use ten prototypes in the 384-dimensional feature space.}
\label{fig:synthetic_prototypes}
\end{figure}

One possible explanation for the gains from prototype alignment over the frozen and finetuned baselines is that the alignment loss simply regularizes the SAR latent space by encouraging class separation, rather than transferring useful information from the optical modality. To test this, we replace the EO derived prototypes with a set of synthetic prototypes. These are artificial class targets that are uniformly separated in the feature space, but are constructed without any EO data, so they do not contain any useful semantic information. Figure~\ref{fig:synthetic_prototypes} shows a schematic 2D example of EO and synthetic prototypes. EO prototypes retain class geometry through their pairwise angles, whereas synthetic prototypes impose uniform separation. If the improvement were only due to spreading classes apart geometrically, synthetic prototype alignment should perform similarly to EO prototype alignment.

Table~\ref{tab:synthetic_results} shows that synthetic prototype alignment reaches only $29.3\%$ accuracy and $0.242$ macro-F1, close to the SAR only finetuning baseline in Table~\ref{tab:main_results} but below the performance of EO prototype alignment. This suggests that the benefit of EO prototype alignment is not simply due to latent space regularization. Instead, the optical prototypes appear to encode class structure that is useful for SAR classification.

\begin{table}[htbp]
\centering
\begin{tabular}{lccc}
\toprule
Method & Acc. (\%) & Macro-F1 \\
\midrule
Synthetic prototype alignment &  $29.3 \pm 2.0$ & 0.242 \\
EO prototype alignment &  $\mathbf{33.3 \pm 1.3}$ & $\mathbf{0.281}$ \\
\bottomrule
\end{tabular}
\vspace{4pt}
\caption{Control experiment comparing synthetic latent space vectors with EO class prototypes.}
\label{tab:synthetic_results}
\end{table}

The structure of the EO prototype space also supports this interpretation. By measuring pairwise angles between the EO prototypes, we find that visually similar vehicle classes lie close together in feature space. For example, the angles are $6.5^\circ$ between \textit{pickup truck} and \textit{sedan}, $6.8^\circ$ between \textit{SUV} and \textit{sedan} and $9.8^\circ$ between \textit{SUV} and \textit{pickup truck}. These classes have similar shape and scale and may therefore be difficult to distinguish. In contrast, more visually distinct classes are farther apart, with an angle of $56.9^\circ$ between \textit{flatbed truck} and \textit{motorcycle}. The small angles among the light vehicle prototypes are consistent with the overlap among sedans, SUVs, vans and pickup trucks in the t-SNE projection in Fig.~\ref{fig:tsne}. Thus, unlike the synthetic prototypes, the EO prototypes not only impose class separation but also preserve class structure derived from the optical features.



\section{DISCUSSION AND FUTURE WORK}
\label{sec:discussion}

The results indicate that optical foundation models can be useful for SAR ATR even when EO imagery is unavailable at inference time and pair identities are not used during training.
Beyond finetuning, the results suggest that the benefit comes from class level transfer from EO prototypes to the SAR representation. The EO prototypes provide class references derived from optical imagery and the SAR encoder learns to align its embeddings to those references while still performing supervised SAR classification.
Although SAR and EO images are formed by different sensing mechanisms, many vehicle classes retain enough shape and scale information for this optical structure to be useful. This is reflected in the t-SNE visualization in Section~\ref{sec:tsne}, where the projections provide qualitative evidence of better separation among some vehicle classes after EO prototype alignment. The remaining overlap among sedans, SUVs, pickup trucks and vans motivates grouping them into a light vehicle superclass. In this evaluation over seven classes, prototype alignment reaches $52.2\%$ accuracy, compared with $41.1\%$ for frozen DINOv3.

The control experiment in Section~\ref{sec:EO_regularization} appears to support this interpretation. If prototype alignment only acted as a generic regularizer, synthetic prototypes with uniformly separated targets should have produced similar gains. Instead, synthetic prototype alignment performs close to simple finetuning and worse than EO prototype alignment. This suggests that the improvement comes not only from encouraging class separation, but from the information encoded in the EO prototypes.

There are also limitations. First, the method still uses class labels in both modalities, so it should be described as a supervised cross-modal transfer method rather than unsupervised domain adaptation.
Second, although the training objective does not use pair identities, experiments are still conducted on a dataset that contains coincident EO/SAR. A stronger test of the proposed setting would build EO prototypes from samples that are disjoint in vehicle instance, location and/or acquisition event from the SAR training data. Third, the absolute classification scores remain low, reflecting the difficulty of the UNICORNv2 dataset, the severe class imbalance and subtle visual differences between vehicle categories in the highly speckled SAR images. Finally, the experiments are limited to civilian vehicle classes, so further validation is needed before drawing conclusions about other SAR ATR use cases, sensing domains or target classes.

Future work should investigate which forms of cross-modal information are useful for transfer. Several MAVIC-C approaches use coincident EO/SAR pairs explicitly during training \cite{ramos2026mavic, liu2025_MAVICEntry}. Their reported performance is broadly comparable, although differences in evaluation protocols prevent direct comparison and make it difficult to isolate the benefit of pairing. A more systematic comparison could determine which shared information and forms of supervision are most useful for transfer, building on recent work on shared information and identifiability in multimodal representation learning \cite{daunhawer2023_identifiability, timilsina2024identifiable}.

The approach could also be extended to settings where substantially more EO than SAR data is available. An extreme case is zero-shot cross-modal transfer\cite{socher13_zero-shot_cross-modal}, where EO examples are available for a new class but no SAR examples are observed for that class. The model would then need to infer how the new EO class maps into the SAR feature space using relationships learned from classes observed in both modalities.



    \section{CONCLUSIONS}
    \label{sec:conclusion}
    
    This work presents a cross-modal transfer approach that uses EO data and optical vision foundation models to improve SAR target recognition. A frozen DINOv3 encoder constructs class level EO prototypes by averaging feature vectors within each class. A separate SAR encoder is adapted using LoRA and trained to classify images while aligning its embeddings to the corresponding EO prototype. Although EO data is used during training, the final model only requires SAR imagery at inference.
    
    On the UNICORNv2 dataset, EO prototype alignment improves over frozen DINOv3, SAR only LoRA finetuning and unpaired MMD alignment. The t-SNE visualization provides qualitative evidence of more separated SAR embeddings after prototype alignment, while the remaining overlap among sedans, SUVs, pickup trucks and vans motivates the light vehicle superclass evaluation, in which prototype alignment also outperforms frozen DINOv3.
    A control experiment with synthetic prototypes suggests that the improvement is not explained only by latent space regularization, but is related to structure encoded in the EO prototypes by the optical foundation model. These results point to a practical route for using large scale optical representation learning to improve SAR ATR without relying on one-to-one EO/SAR correspondences in the training objective.
    
    \appendix    

  \section{SIMULATION DETAILS}
    \label{sec:appendix}
    Full configurations, implementation details and code used to produce the main results in Table~\ref{tab:main_results} are available on GitHub at \url{https://github.com/luhirsch/eo-to-sar-prototype-alignment}. Method settings are summarized in Table~\ref{tab:implementation_details}.
    \begin{table}[htbp]
    \centering
    \begin{tabular}{@{}lccc@{}}
    \toprule
    Method & Learning rate & Alignment & Weight $\lambda$ \\
    \midrule
    Frozen DINOv3
        & $1\times10^{-3}$ & -- & -- \\
    SAR only LoRA
        & $4\times10^{-6}$ & -- & -- \\
    Unpaired MMD
        & $8\times10^{-6}$ & MMD & 0.45 \\
    EO prototypes
        & $8\times10^{-6}$ & Cosine distance & 0.60 \\
    \bottomrule
    \end{tabular}
    \vspace{4pt}
    \caption{Training hyperparameters for each method.}
    \label{tab:implementation_details}
    \end{table}
    
    All experiments used the pretrained DINOv3 ViT-S/16+ backbone, which outputs features of dimension $d=384$. EO and SAR images were resized to $128\times128$ and normalized using ImageNet statistics. No additional augmentations were applied. All models were trained with a batch size of 256, the AdamW optimizer, cosine learning rate scheduling and weighted sampling with replacement using inverse class frequencies. For MMD alignment, SAR and EO data were sampled independently to prevent paired images showing up in the same minibatch. Models were trained for at most 25 epochs, retaining the checkpoint with the highest validation macro-F1. Most runs reached their best validation performance within the first 10 epochs.
    
    For the frozen baseline, only a linear classification head was trained. All other methods trained the linear head and LoRA adapters applied to the query ($Q$), key ($K$) and value ($V$) projection matrices in each transformer block of the DINOv3 backbone. The LoRA parameters were set to rank $r=28$, scaling $\alpha = 56$ and dropout 0.05.
    
    For the full loss shown in Eq. \eqref{eq:full_loss}, $\lambda$ was set to 0.60 for EO prototype alignment. For MMD, $\mathcal{L}_{proto}$ was replaced by an MMD loss $\mathcal{L}_{MMD}$ computed using RBF kernels and $\lambda$ was set to 0.45. The $\lambda$ values for both methods were selected after sweeping $\lambda$ from $0.1$ to $0.8$ in increments of $0.05$.

    \acknowledgments 
     
    This work was supported by the Engineering and Physical Sciences Research Council (EPSRC) under Grant number EP/X025365/1, in collaboration with Leonardo, as part of the 'Smart Products Made Smarter' project.
    
    \bibliography{report} 
    \bibliographystyle{spiebib} 

    \end{document}